# Misgendering and Assuming Gender in Machine Translation when Working with Low-Resource Languages

*Sourojit Ghosh, University of Washington, Seattle, USA*
*Srishti Chatterjee, University of Melbourne, Australia*

## Abstract

This chapter focuses on gender-related errors in machine translation (MT) in the context of low-resource languages. We begin by explaining what low-resource languages are, examining the inseparable social and computational factors that create such linguistic hierarchies. We demonstrate through a case study of our mother tongue Bengali, a global language spoken by almost 300 million people but still classified as low-resource, how gender is assumed and inferred in translations to and from the high(est)-resource English when no such information is provided in source texts. We discuss the postcolonial and societal impacts of such errors leading to linguistic erasure and representational harms, and conclude by discussing potential solutions towards uplifting languages by providing them more agency in MT conversations.

## 1. Introduction

Translation has been widely acknowledged as a power practice. Either by closing or opening doors, translators are considered as gatekeepers that may exclude or include a person, a value system, a language community, or an entire culture (Venuti 1992). Translation bridges the gap between dominant and minoritized languages within a global context (Cronin 2003), in bilingual speakers/communities (Edwards 2012), and in superdiverse societies (Drugan and Kredens 2018). It is a tool of equity in access to justice, health services, education, and other global contexts (Bancroft 2015; Khelifa, Amano, and Nuñez 2022; Wolfenden, Cross, and Henry 2017). Translation is not simply a reproduction of information, but rather a production of knowledge by a human translator or a machine agent (Andone 2002), which embeds the translator's or programmer's political and ideological views. Translation can thus be an effective tool for resistance and revolution (Niranjana 2023; Tymoczko 2010), but alternatively a powerful means of normalizing an opinion and knowledge among people. When translation is handed over to algorithmic systems, harms may ensue (Blodgett et al. 2020). Machine translation (MT) thus holds the power to uphold and enforce social power dynamics on a regular basis, and a strong example of that is the flattening and homogenizing gender and ethnic diversities of languages, most commonly those which suffered colonial oppression.

In this chapter, we explore the sociotechnical implications of MT in the context of gender in translation to/from or between low-resource languages, specifically our mother tongue of Bengali. We do not present novel empirical data, but instead build upon prior work and provide discursive commentary from our identities as native speakers of a rich language which still is assigned the moniker low-resource. We also apply our perspectives of being regular users of machine translation both in official and personal use cases, and linguistic researchers who study the sociotechnical impacts of erroneous MT between high- and low-resource languages.

## 2. Low-Resource Languages

What makes a language high- or low-resource is a combination of several factors, a concept that this chapter addresses by highlighting key aspects. In the early 1950s, Natural Language Processing (NLP) techniques relied on simple rule-based approaches based on human-dictated lists of rules on how to interpret, manipulate, and evaluate one or more symbols fed to computers, which would then execute these instructions on new inputs (Winograd 1971). This approach was unscalable, and the mid 90s brought statistical approaches and machine learning (ML) techniques, with processes such as decision trees, parts-of-speech tagging, and stemming and *n*-gram approaches, to name a few. Modern-day NLP uses such features in combination with robust computational capabilities, enabling the handling of many terabytes of data.

Consequently, there is a growing impetus to amass data on such scales to fuel the ever-expanding demands of modern NLP. Data collection efforts, especially for MT purposes, consequently began looking for pairs of source-and-translated texts with which to train models, with the most common early candidate being the Bible (Costa-jussà et al. 2022). This made sense, since the Bible was not only primarily circulated in Western languages such as English, French, Spanish, and Portuguese, but also in languages of the Indian subcontinent, Latin America, Africa, and Oceania (McCarthy et al. 2020). Here, it is important to remember why and how the Bible achieved such a status as one of the most translated texts.

Any conversation around the hierarchies of languages would be incomplete without considering various sociopolitical forces such as imperialism and colonialism that, unintentionally or (more often) intentionally, shaped their positions along linguistic ladders. Colonial efforts of the UK, Spain, Portugal, and France, among others, and their quests for global domination saw them expand their respective empires into Asia, Africa, and America. As they enslaved the indigenous people through economic or military means, colonizers also imposed their languages, often with the intention of erasing local languages (Ravishankar 2020). The erasure of local languages and their replacement with colonizers' tongues rendered people so beholden to the language that liberation was not considered feasible. Parallelly, as colonizers brought Christianity with them, a need arose to establish versions of the Bible in the local languages of the colonies. With the British colonial empire being the largest, English grew into the global lingua franca, a phrase embedding colonialism itself because it translates into 'language of the Franks,' a collective referring to Western Europeans, yet the phrase is now used globally.

Therefore, English has become one of the most-spoken languages, with Spanish, French, and Portuguese joining in. Over time, such languages grew to become *high-resource* languages, or those that have strong coverage and significant representation in global datasets (Cieri et al. 2016), whereas other languages that do not have as strong coverage are termed *low-resource*. Linguistic resource is a function of the volume of text in a language available on the Internet which can be used to train NLP models, and consequently, the reasons for the availability of such volumes of text or lack thereof are what make languages low- or high-resource. We address a few of those below.

In the context of collecting and forming global datasets, an important factor is the advent and spread of the Internet in various parts of the world. Since the Internet was started and popularized in North America and Europe, their users were the first to participate on the Internet to produce content. One reason for the rise of European languages in the hierarchy is the Europarl (2005) dataset, compiled from 11 languages used within the European Parliament. But Internet availability also created hierarchical differences between non-European languages. Internet availability simultaneously created hierarchical differences between non-European languages. Consider the case of China and India. These launched publicly-available Internet access within a year of each other (China in 1994 and India in 1995) and yet, China's 420 million Internet users is over 5 times that of India's 81 million (Prakash 2018). There are several suspected and plausible reasons for this gulf, such as differences in literacy levels, in median incomes and ability to afford Internet access, higher number of mobile phones and Internet-capable devices manufactured and used in China.

Another reason that can equalize the population difference is once again a political one: linguistic oppression and standardization. Staying on the aforementioned example, Chinese is the most predominant language in its country due to long histories of internal linguistic oppression and state-driven efforts to standardize it by relegating variations to dialect statuses (Hartman 2020). On the other hand, India boasts 121 unique languages and their dialects (PTI 2018). There are also significant differences between Indian languages, with languages such as Hindi and Bengali rooted in Sanskrit whereas those like Tamil and Telugu descend from Dravidian languages. Therefore, there could possibly exist a far higher number of people who speak Chinese fluently as first-language speakers, over any one Indian language, with the possible exception of Hindi.

The number of people who speak a certain language has nothing to do with its status as low- or high-resource, and there exist several languages, such as Swahili, Urdu, and Thai, spoken by millions of people but still considered low-resource because of the disproportionate amount of text available. In this section, we covered a few reasons why these amounts vary, although these are but contributors to a complex landscape of global and local language hierarchies, including linguistic ideologies and specific intergroup conflicts (see Castelló-Cogollos and Monzó-Nebot 2023).

## 3. Case Study: The Bengali Language

### 3.1 Author Positionality

We present a case study of our mother tongue: Bengali. The authors are native speakers of Bengali, in the Rahri and Bangali dialects, having been born and raised in Kolkata, the capital city of the Indian state of West Bengal where Bengali is spoken as the official language by almost 3 million people (Mukherjee 2014). The first author's research experience relates to MT for Bengali, with first-authored original work in this space. The second author's research experience is in gender and algorithmic justice, and governance. Their positionality is also influenced by their lived experience of gender diversity, as a non-binary person.

### 3.2 The Bengali language

Bengali is within the top 10 most-spoken languages in the world, with around 300 million native speakers (Ghosh and Caliskan 2023). It is primarily spoken as the official language of Bangladesh and the state language of Bengal in India, two regions which used to be united before the 1905 colonial partition of erstwhile-Bengal by British colonizers. Bengali is recognized as one state language in the states of Assam and Tripura (Das, Østerlund, and Semaan 2021), and the country of Sierra Leone (Ghosh and Caliskan 2023). Bengali has several dialects, with Rahri being the most common and closest to the governmental Standard Colloquial Bengali used for official purposes (Ghosh and Caliskan 2023).

Bengali occupies a special place in the global language family, because the International Mother Language Day (ভাষা দিবস, pronounced 'Bhasha Dibosh') is celebrated annually on 21st February. This recurrence was first declared by UNESCO in 1999 and the 21st February commemorates the Language Movement of then-East Pakistan (presently, Bangladesh) when students of Dhaka University led rallies in 1952 to champion the cause of Bengali being one of the official languages and were killed at the hands of then-West Pakistan (presently, Pakistan) police. Their deaths evoked public outrage and led to the official constitutional status of Bengali in then-East Pakistan, also subsequently the liberation and establishment of the sovereign Bangladesh in 1971. 21st February is celebrated in Bangladesh and Bengal with Abdul Gaffar Choudhury's song একুশের গান (pronounced 'Ekusher Gaan', meaning '21st's song'), popularly known as আমার ভাইয়ের রক্তে রাঙানো একুশে ফেব্রুয়ারি (pronounced 'amar bhaiyer rokte rangano ekushe February', meaning 'my brother's blood colors the 21st of February').

The Bengali language has a strong literary canon, with the work of Asia's first Nobel laureate in literature, Rabindranath Thakur (we refuse to use the popularly anglicized *Tagore*, offering resistance to the idea that non-English names need to use their colonial anglicizations). Thakur routinely wrote and translated his work between English and Bengali, and it is for his English translation of *Gitanjali* as popularized by poet WB Yeats that he won his Nobel prize. Aside from a strong history in its own culture, Bengali is one of the first colonial languages to be translated into English, and boasts a large volume of English–Bengali text.

Despite Bengali's rich heritage and large speaker base, past research (Das, Guha, and Semaan 2023; Ghosh and Caliskan 2023) has shown that MT from Bengali into English and vice versa still is still rife with errors, specifically in the context of gender. We will elaborate on this in the next subsection. First, a brief explanation of how gender works in Bengali is provided.

## 3.3 Gender in Bengali, and Gender-related Issues in English ↔ Bengali Machine Translation

Unlike English, which has gendered third-person pronouns such as *he/him/his* or *she/her/hers*, Bengali uses gender neutral third-person pronouns – সে [*shey*], ও [*o*] and তিনি [*teeni*]. There are other pronouns such as এ [*ey*, as in *whey*], উনি [*uuni*], ইনি [*ini*], among others. All of these correspond to the English singular gender-neutral *they/them/theirs*.

Despite lacking gendered pronouns, Bengali still contains binary gender, and often uses variations of the same word to denote genders (Ghosh and Caliskan 2023). For example, সিংহ refers to lions, and is modified to সিংহী for lionesses. Similarly, the word শিক্ষক refers to male teachers, and is modified to শিক্ষিকা for female teachers. It is, however, important to recognize a growing movement within Bengal to use the default unmarked word (in this case, শিক্ষক) as the gender-neutral version and avoid continually invoking the binary embedded within words (Ghosh and Caliskan 2023). As is the case with many global languages, male versions of Bengali words usually perform as the unmarked default, and the marking is added to generate a female version (Wolfe and Caliskan 2022). In some cases, different words are used for male-female pairs, such as রাজা, which translates to 'king' and রাণী to 'queen.'

In the context of gender errors in English ↔ Bengali translation, Ghosh and Caliskan's (2023) study on one of the state-of-the-art large language models (GPT-3) is relevant. They created single-occupation prompts with 50 Bengali sentences constructed as সে একজন ___ (meaning, 'they are a ___') and blanks filled with the Bengali word for the top 50 most common US occupations according to the Bureau of Labor Statistics (2022). While they expected translations into English to use the gender-neutral *they* pronouns, GPT-3 inferred and inserted binary gender into translations based on occupations, where doctors, inspectors and physicians were assigned the male *he*, whereas nurses, receptionists and assistants were assigned the female *she*. When translating 50 English prompts constructed as 'They are a ___' and filling in the blanks with the English occupation titles as above, GPT-3 was completely unable to identify *they* to be gender-neutral and singular, and incorrectly translated it to the plural[1] Bengali pronoun তারা. Furthermore, they demonstrate how the inferred gender in Bengali to English translations is predicted more strongly based on the action mentioned rather than the occupation in sentences containing both. Actions such as waking up in the morning and cooking food or cleaning the house would return *she* pronouns when translated into English, overwriting male pronouns previously seen for occupations such as doctors and engineers. Ghosh and Caliskan extended these findings by forming similar prompts and testing them across five other languages – Farsi, Malay, Tagalog, Thai, and Turkish – to demonstrate that the problem is common across other low-resource languages.

There exist other epistemic experiences of gender errors in English ↔ Bengali translation (Ghosh and Sengupta 2021). From their own lived experiences, the authors have seen this first-hand, both in Google Translate and ChatGPT, and we are sure our readers who use such services for languages that deal with referential gender differently than English can also relate. Outside of Bengali, both research (Caliskan 2023; Cho et al. 2019; Prates, Avelar, and Lamb 2020) and epistemic experiences (Morse 2017; Ullmann and Saunders 2021), highlights the largely-unsolved problem of gender issues in MT to and from low-resource languages. Such issues can be attributed to a variety of reasons, such as the absence of contextual information in source texts from which the machine translator can infer gender, quality and accuracy issues within the corpora of training datasets, traditionally sexist word associations embedded within training data (for example, associating 'nurse' with 'she'), among other factors. Though such issues occur in high-resource languages, such as Spanish and Italian (Rescigno et al. 2020; Stanovsky, Smith, and Zettlemoyer 2019), that high-resource languages traditionally deal with referential gender similarly due to shared roots widens the gulf between low- and high-resource languages (Ghosh and Caliskan 2023; Ryu 2017). These errors should not merely be considered simple mistakes, but rather be considered for the broader sociotechnical implications they bring (Norman 2013), which we explore in the next section.

## 4. Representational Harms Caused by Gender-Errors

With MT systems being widely utilized in global contexts and different cultures while carrying such biases and errors, recognition of such biases have mostly been met with a drive to increase accuracy and removal of incorrect translations. Surveys to take stock of MT practices in trying to rectify such errors have largely been technosolutionistic, simply considering technical problems that have technical solutions. Such an approach, both in this specific case and generally across NLP/ML, misses the opportunity to situate them in the societal context within which the problems occur (Birhane 2021).

The societal impacts of NLP/ML cause, among other types of harm, *representational harms* — harms caused when people or groups are stereotyped and stigmatized (Barocas et al. 2017). With gender bias and errors in MT, representational harms can occur through two different ways, stereotyping and underrepresentation, and the ways in which they manifest are often intrinsically intertwined.

One of the strongest ways in which stereotyping and underrepresentation manifests is when, as with the examples in section 3.3, language translation associates occupations and social roles with (binary) genders, such as labeling *nurses* as female and *engineers* as men in a large number of cases, both in Bengali and in other languages (Ciora, Iren, and Alikhani 2021; Ghosh and Caliskan 2023; Fitria 2021; Prates, Avelar, and Lamb 2020). Though it is generally understood that such outcomes come from biases in training data and word-embeddings which contain historical human biases along similar lines (Caliskan, Bryson, and Narayanan 2017; Guo and Caliskan 2021), the field seems to not be closer to solving this problem. This is an underrepresentation problem which can manifest, to coin a possible situation as an example, when a Bengali novel which writes a female character as a doctor might run the risk of seeing that translated to a male doctor in English if the MT system does not perceive additional context as female. This is problematic for two reasons: it feeds into the culture of feminine entities being misrepresented as male (Savoldi et al. 2021), and puts the onus on the author to ensure they convince their non-Bengali speaking readers that the doctor is not male.

Stereotyping and underrepresentation of traditionally marginalized genders occurs when MT systems *flatten* the identities and communicative repertoires of speakers of a language, erasing the identities of women and non-binary people into the default male identity (Hovy, Bianchi, and Fornaciari 2020; Savoldi et al. 2021), especially when the language used to speak *about* such communities primarily uses non-gendered pronouns. In Bengali, third-person pronouns are not gendered and referring to people of binary and non-binary genders in third person is not a problem, but translating such pronouns and sentences about people into English forces the model to assign a (binary) gender, and leads to the linguistic erasure of people of nonbinary genders who are originally depicted correctly. Not only does flattening occur with respect to gender in MT, but there is also an ethnolinguistic flattening across the language

hierarchy — that many languages do not structurally reproduce gender through their third-person pronouns is flattened into the binary assumptions of the English language, with the identity of non-binary people and women being subsumed by the male default and, quite literally, lost in translation. This results in languages that do not have similar gender rules to English/the target language suffering more, and arbitrarily so. Genders are flattened into a default male and, consequently, the diversity of gender and how differently several languages handle them are flattened into the colonial binary. Over time, such translation errors could and do become solidified in MT systems and outputs, and with them, their ability to cause representational harms. This can especially be true without native speakers' context of the morphological, syntactical, and historical-material contexts carried in languages with rich (and praxiologically anti-colonial) history and literary canon, as is the case with Bengali.

Concurrently, gender-related errors, both in low-resource languages and beyond, are an expression of the power enacted by MT in their ability to cause representational harms. That a language as morphologically rich and diverse as Bengali (Roy 2010) suffers thus is an extension of this power, because incorrect translations might persist in user generation of documents which might end up on the Internet, continuing to associate সে with 'he/she' and then be used as training data for MT systems to further solidify their incorrect translations. The linguistic flattening of Bengali is not simply a *mistake*; it is an extension of stereotypes about genders and ethnolinguistically diverse cultures and communities ultimately working to homogenize and assimilate them into cis-heteronormative, colonial norms and traditions. To recognize these errors as so much more than simple mistakes is critical for the path forward. Acknowledging the deep colonial history behind them, the erasure of languages, histories, and peoples in the flattening of gender with the urge to drive towards the unmarked defaults (Trubetzkoy 1969), and the representational harm caused to people with identities beyond the default (Jacobs et al. 2020), is a crucial step towards achieving stronger representation for Bengali, other low-resource languages, and languages in general. In the next section, we provide some more steps towards a way forward.

## 5. Ways Forward

### 5.1 Gather More Data, but Mindfully

Similar to why some languages are low-resource as opposed to others, the first solution to the problem of gender errors in translation seems to be to gather more data. Such a solution seems obvious, but it begs the question: if it were such a simple idea, why has it not happened yet? This is because simply gathering more data is non-trivial and requires a sensitive understanding of the various factors that have resulted in a paucity of data in currently low-resource languages, some of which are pointed out in section 2, and are similar to the reasons for a lack of high-accuracy gender-translation data.

The problem of not having enough data is "a sociotechnical failure, where 'better' data is difficult to achieve due to the various social constraints designed to favor languages that are already high-resource" (Ghosh and Caliskan 2023) This extends human-centered machine learning (Chancellor 2023), calling upon researchers to consider that their failures and errors are likely not only restricted to technical issues with code and data, but also due to social factors. Here, an important social factor that contributes to the difficulty of simply collecting data could be the unwillingness of speakers of certain languages to participate in data collection efforts depending on who is collecting the data, due to evidence of past trauma that their communities have gone through at the hands of such people. Any data collection efforts must be conducted with appropriate sensitivity, and a strong understanding of the dynamics and cultures of the people who would be asked to provide the data (see Cronin 2020; and Thorat 2021 on data extractivism and colonialism). This is especially true for extremely-low resource languages or endangered languages, for which finding high-quality data might be an altogether different challenge. Efforts could also be made to engage experienced translators in creating datasets that do not make the aforementioned gender errors, by hand-translating texts between languages that handle referential gender differently.

## 5.2 Engage Local Community Partners and First-Language Speakers

As Ghosh and Caliskan (2023) write, a human-centered approach towards MT and upliftment of low-resource languages must involve active participation of the community members and first-language speakers. Such people are the ones who have the strongest connection to their language, the relevant lived experience of the various cases in which the language and its dialects are used and would be the experts of the usage because they are the most likely users of a tool that can support their language well. For a roadmap on how to do this effectively, we offer the work of Costa-jussà et al. (2022) and the FLORES-200 dataset, which provides coverage of over 200 low-resource languages and data collection procedures in such methods.

Any work on engaging local community partners and first-language speakers must also understand that not all low-resource languages are at the same level. Despite researchers and NLP practitioners clumping them all together in one umbrella term, every individual language has different rules in the context of grammatical gender (and other factors) and is at a very different place in terms of the number of people who speak it, the varying levels of fluency, and where it is on its linguistic journey. While languages such as Bengali are widely-spoken across multiple countries, Indigenous languages are continually facing erasure and extinction due to decreasing numbers of first-language speakers in the face of growing pressures to adopt state languages (often the language of their colonizers) and speakers of such languages are often dedicating their entire lives to the preservation and celebration of the languages and the rich cultural heritage that face colonial erasure. To believe that the same strategies would fix gender-related errors in translation and increase linguistic representation of a language like Bengali and an Indigenous language which is fighting for survival is not only incorrect, but would demonstrate a deeply insensitive and flawed understanding of sociopolitical conditions (Howard, Ricoy, and Ciudad 2018; Tonja et al. 2023).

Even with best intentions and attempts to work with community partners, we believe that it is unfair for MT/NLP researchers and practitioners to practice epistemic extractivism (Grosfoguel 2020), simply gathering data from speakers of low-resource languages without contributing in tangible ways. For instance, in the aforementioned context of Indigenous languages, researchers and those who are able to do so should consider finding out opportunities to donate to local Immersion Schools and language-learning camps to support the efforts of the people working tirelessly to keep their languages alive. For other languages and other cultures, researchers must do their own groundwork to find out the best possible ways to give back to the communities they are working with, the best practice towards which is to often consult the communities themselves.

The task of resolving gender issues in translation between low-resource languages or to/from low- and high-resource languages is not simply one that can be solved by going to speakers of such languages and gathering data. Rather, it is a complex sociotechnical process that requires a keen understanding of local community cultures and a willingness to give back to such communities.

## 5.3 Supporting Interdisciplinary, Multidisciplinary, and Multilingual Research

While writing this piece, we came across an overwhelmingly high number of articles that addressed biases in the fields of MT and Translation Studies, that had excluded papers which were written in, or had significant components written in languages other than English. Though anglocentrism is not unique to this field, we feel that it is particularly significant to call out, even as we include ourselves in the group of researchers writing primarily in English. To move towards fairer linguistic representation, the involvement of community partners and first-language speakers should be pedagogically included in research, review, and policymaking, with their voices being represented in the work in their own languages. The research should not be purely technical, or socio-cultural, with a belief that the answer exists in stronger NLP work instead establish that the solution is a sociotechnical one requiring interdisciplinary work (Blodgett et al. 2020).

Although there exists commonalities and correspondences between languages, each language is unique, and they should not collectively be homogenized to the high(est) resourced English (Bußmann 2003). In MT models, this research involves ethnically, culturally, and gender-informed critical reflections of how cultural and gendered identities are "intended and ascribed" (Larson 2017). Such research and implementation should include multidisciplinary experts on socio-cultural identity, technical knowledge, as well as epistemic experiences of people of diverse genders and ethnicities.

# 6. Conclusion

In this chapter, we explored gender-related errors in MT in the context of low-resource languages. Through an examination of prior research and epistemic experiences, we show how prominent MT services such as Google Translate and ChatGPT mishandle and erroneously interpret grammatical gender in the context of English ↔ Bengali translations. We provide discursive commentary on the representational harms caused by such errors as products of colonialism, and why they have far broader implications than simply being annoying issues to be fixed and moved on from. We conclude with imagining ways forward, encompassing closer collaboration with individual linguistic communities and a willingness to support them in valuable ways.

There is a lot of space for future work and several open questions remain. Future researchers could adopt our recommendations listed in section 5 and apply them in their work, documenting how they work in practice and any potential challenges that need to be addressed. Another extension could be to consider work with specific communities and their endangered or extremely-low resource languages, and how local community support and expertise could be centered in meaningful ways that result in uplifting the language and building tools to support the people who speak it.

## Notes

1. Consider the following two English sentences: *They are a doctor*, and *They are doctors*. The Bengali word তারা would be an appropriate translation of the *they* in the second sentence, and not in the first. However, GPT-3 would translate the *they* in the first sentence also to তারা, when the correct translation would be a singular pronoun such as সে, ও, or তিনি .